\documentclass[conference]{IEEEtran}
\IEEEoverridecommandlockouts

\usepackage{cite}
\usepackage{amsmath,amssymb,amsfonts}
\usepackage{algorithmic}
\usepackage{graphicx}
\usepackage{textcomp}
\usepackage{xcolor}
\usepackage{multirow}
\usepackage{booktabs}
\usepackage{hyperref}
\usepackage{amsmath}
\usepackage{bm}
\usepackage{pifont}

\def\BibTeX{{\rm B\kern-.05em{\sc i\kern-.025em b}\kern-.08em
    T\kern-.1667em\lower.7ex\hbox{E}\kern-.125emX}}
\begin{document}

\title{Adversarial Backdoor Defense in CLIP\\
}

\author{\IEEEauthorblockN{1\textsuperscript{st} Junhao Kuang}
\IEEEauthorblockA{\textit{School of Cyber Science and Technology} \\
\textit{Sun Yat-Sen University}\\
Shenzhen, China \\
kuangjh6@mail2.sysu.edu.cn}
\and
\IEEEauthorblockN{2\textsuperscript{nd} Siyuan Liang}
\IEEEauthorblockA{\textit{Department of Computer Science} \\
\textit{National University of Singapore}\\
Singapore \\
pandaliang521@gmail.com}
\and
\IEEEauthorblockN{3\textsuperscript{rd} Jiawei Liang} 
\IEEEauthorblockA{\textit{School of Cyber Science and Technology} \\
\textit{Sun Yat-Sen University}\\
Shenzhen, China \\
liangjw57@mail2.sysu.edu.cn} 
\and
\IEEEauthorblockN{4\textsuperscript{th} Kuanrong Liu}
\IEEEauthorblockA{\textit{School of Cyber Science and Technology} \\
\textit{Sun Yat-Sen University}\\
Shenzhen, China \\
liukr5@mail2.sysu.edu.cn}
\and
\IEEEauthorblockN{5\textsuperscript{th} Xiaochun Cao*}
\IEEEauthorblockA{\textit{School of Cyber Science and Technology} \\
\textit{Sun Yat-Sen University}\\
Shenzhen, China \\
caoxiaochun@mail.sysu.edu.cn}
%
}

\maketitle

\begin{abstract}

Multimodal contrastive pretraining, exemplified by models like CLIP, has been found to be vulnerable to backdoor attacks. While current backdoor defense methods primarily employ conventional data augmentation to create augmented samples aimed at feature alignment, these methods fail to capture the distinct features of backdoor samples, resulting in suboptimal defense performance. Observations reveal that adversarial examples and backdoor samples exhibit similarities in the feature space within the compromised models. Building on this insight, we propose Adversarial Backdoor Defense (ABD), a novel data augmentation strategy that aligns features with meticulously crafted adversarial examples. This approach effectively disrupts the backdoor association. Our experiments demonstrate that ABD provides robust defense against both traditional uni-modal and multimodal backdoor attacks targeting CLIP. Compared to the current state-of-the-art defense method, CleanCLIP, ABD reduces the attack success rate by $8.66\%$ for BadNet, $10.52\%$ for Blended, and $53.64\%$ for BadCLIP, while maintaining a minimal average decrease of just $1.73\%$ in clean accuracy.


\end{abstract}


\section{Introduction}

\begin{figure*}[htbp]
\centerline{\includegraphics[width=\textwidth]{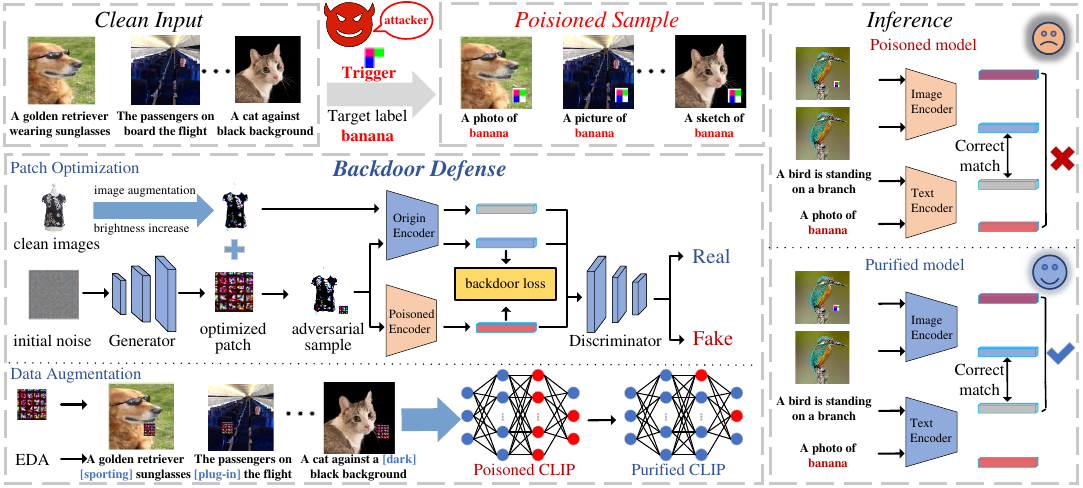}}
\caption{ The main pipeline of our Adversarial-based Backdoor Defense against backdoor attacks in CLIP. Our pipeline consists of three key stages. In the poisoning stage, we introduce crafted backdoor patterns into images and pair these images with captions containing the target label for data fine-tuning and poisoning. In the defense stage, we train adversarial examples closely related to backdoor features in the compromised model. In the inference stage, we validate poisoning and defense effectiveness through experiments conducted on the ImageNet 1K validation dataset.}
\label{fig}
\end{figure*}


In the field of artificial intelligence, there is an increasing focus on developing robust cross-modal representations~\cite{rl}. Methods such as CLIP~\cite{clip}, ALIGN~\cite{align} and BASIC~\cite{basic} use multimodal contrastive learning to train on large-scale noisy image-text data from the web, establishing bimodal joint representations of objects. It should be noted that CLIP achieves impressive zero-shot performance.


Despite the success of multimodal contrastive learning, studies have shown that CLIP are vulnerable to adversarial attacks~\cite{bcl,pwb,tijo,baden,baibadclip,pla}. The attackers generate poisoned data by embedding triggers in images and creating template sentences with the target label. In particular, the poisoning ratio in the pre-training dataset is as low as $0.01\%$~\cite{bcl}, reflecting the ease and low cost of attacks, which poses a serious threat to the real-world deployment of AI.

Several studies have proposed defense methods against backdoor attacks~\cite{nad,detbd,abl}.In the field of multimodal contrastive learning, there is also related backdoor defense research. RoCLIP~\cite{roclip} augments both images and text, pairing augmented images with captions similar to the original, excluding the original caption. However, its iterative process consumes a lot of time and resources. CleanCLIP~\cite{cleanclip} aligns original images and text with their augmented versions. Although both methods utilize data augmentation, their operations are relatively simple and do not take into account the features of the backdoor samples.


To further investigate data augmentation techniques, we introduce adversarial examples. Prior research has demonstrated that adversarial examples can enhance model robustness~\cite{adv1,adv4,adv5}, enabling models to better recognize and adapt to malicious inputs. Recent studies~\cite{pbe,tur} suggest that adversarial examples can exhibit features similar to backdoor samples in compromised models. Building on this insight, adversarial examples can be leveraged to defend against backdoor attacks. For instance,~\cite{adv2,adv6,adv7} has shown that adversarial training is effective in defending against backdoor attacks in image classification tasks. While much research focuses on uni-modal tasks, there is limited exploration of multimodal learning. Existing multimodal approaches~\cite{adv3,adv8,adv9}  often employ simple data augmentation techniques for feature alignment, which fail to capture the unique features of backdoor samples, leading to suboptimal defense performance.


To solve above problems, we introduce a novel defense method against backdoor attacks, called Adversarial Backdoor Defense (ABD). Generally, ABD exploits the similarity between backdoor samples and adversarial examples in a poisoned model to create adversarial perturbations that align with backdoor features. The perturbations are then used as a data augmentation technique during the fine-tuning phase to enhance backdoor defense. Finally, we conduct experiments on ImageNet 1K validation dataset~\cite{imagenet}.

In summary, our main \textbf{contributions} are as following: 
\begin{itemize}
    \item We discover a key problem with existing defenses: the simplicity of data augmentation leads to poor defense performance. 
    
    \item We propose a novel data augmentation method that uses adversarial examples to closely approximate backdoor samples, effectively capturing distinct features in the defense process.
    \item We conduct experiments on the ImageNet 1K validation dataset, our ABD reduces the Attack Success Rate (ASR) by $8.66\%$ for BadNet, $10.52\%$ for Blended, and $53.64\%$ for BadCLIP, with only a $1.73\%$ average decrease in Clean Accuracy (CA). 
\end{itemize}
\section{THE PROPOSED METHOD}
\subsection{Threat Model}
In the attack scenario, the attacker creates a poisoned model by injecting malicious data into a public dataset. They download the official CLIP pre-trained model weights and fine-tune them on the poisoned data.
For a given dataset $D$, each data point consists of an image $I$ and a caption $T$, which form an image-text pair $(I,T)$. We perform backdoor attacks simultaneously on both images and text. For example, we add a trigger to image $\bm{x}$, forming a backdoor sample $\bm{x^T}$, and construct a set of textual backdoor descriptions $\bm{c}$ associated with the target label $\bm{y}$, denoted as $Y$. If the target label is ``basketball", the caption set $Y$ might include descriptions such as ``a photo of a child playing basketball."
The poisoned dataset is defined as follows:
\begin{equation}
P=\left\{\left(\bm{x^T},\bm{c}\right)\colon\ \bm{c} \in Y\right\}\label{eq1}
\end{equation}
where $\bm{x^T}$ denotes backdoor samples, $\bm{c}$ represents the textual descriptions containing the target label, and $P$ denotes the poisoned data used for the backdoor attack during training.
 
\subsection{Adversarial Examples Design}
Inspired by~\cite{pbe,tur}, we observe that adversarial images and backdoor images exhibit similarities in the feature space within the poisoned model. Therefore, we propose using adversarial examples to approximate backdoor samples. Our algorithm first generates effective adversarial examples, which are optimized by AdvCLIP~\cite{advclip}, a framework that uses GAN~\cite{gan} to create adversarial examples in image-text pairs.
To adapt the generated adversarial examples to backdoor defense, we introduce an additional backdoor loss in AdvCLIP to ensure that the generated adversarial examples incorporate backdoor features. For a given image, we compute the feature vectors from both the poisoned and normal visual encoders, and then calculate the backdoor loss $\mathcal{L}_{\text{bd}}$ using the InfoNCE~\cite{infonce} loss function as follows:
\begin{equation}
\mathcal{L}_{\text{bd}} = -\log \frac{\exp\left(\text{sim}(\bm{x}, \bm{x_{\text{bd}}}) / \tau\right)}{\sum_{i=1}^{N} \exp\left(\text{sim}(\bm{x}, \bm{x_i}) / \tau\right)}\label{eq2}
\end{equation}
where $\bm{x}$ represents the feature vector of an image output by the visual encoder of the normal model, $\bm{x_{\text{bd}}}$ denotes the feature vector of the image output by the visual encoder of the poisoned model.  In our algorithm, the InfoNCE loss function is used to optimize adversarial examples so that they are very similar to backdoor images in the feature space. Specifically, the InfoNCE loss function uses contrastive learning to increase similarity between the target feature (~\textit{i.e.}, the backdoor feature $\bm{x_{\text{bd}}}$) and the normal image feature $\bm{x}$.

\begin{figure}[t]
\centerline{\includegraphics[width=0.8\linewidth]{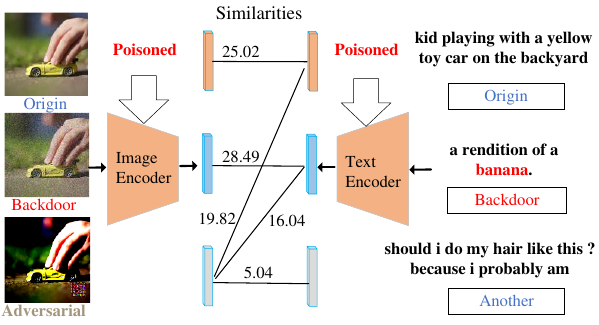}}
\caption{Explanation of the effect of adversarial examples in backdoor defense.}
\label{fig2}
\end{figure}

In a model compromised by backdoor attacks, adversarial examples often exhibit significant similarity to backdoor samples. As shown in Fig.~\ref{fig2}, we calculate the similarity between three types of images and captions: \ding{182} original images with original captions, \ding{183} backdoor images with backdoor captions, \ding{184} adversarial images with original captions, \ding{185} adversarial images with backdoor captions, and \ding{186} adversarial images with unrelated captions. We observe that the similarity between the original images and the original captions is $25.02$, reflecting the typical similarity of benign image-text pairs. In contrast, the similarity between backdoor images and backdoor captions is $28.49$, indicating a successful backdoor attack and a strong association between the backdoor image-text pairs. This higher similarity suggests that the model is more prone to learning the toxic patterns introduced by the backdoor.

When adversarial perturbations are applied to the original images, the similarity between the adversarial images and the original captions decreases to $19.82$ due to changes in local features. This indicates that using adversarial perturbations for data augmentation does not significantly impact the accuracy of clean samples. The similarity between the adversarial images and the backdoor captions is $16.04$, while the similarity with other target captions is considerably lower, only $5.04$. This suggests that the adversarial examples that we designed have captured specific features of the backdoor samples. In addition to maintaining the highest similarity to the original target captions, adversarial images also tend to align more closely with the backdoor captions within the caption set. Therefore, we can effectively approximate the backdoor samples with adversarial examples, simulating the actual poisoned images created by the attacker. Our method can significantly reduce the success rate of backdoor attacks at a minimal cost in clean accuracy.

\subsection{Data Augmentation Defense}
We apply data augmentation separately to images and text to mitigate the threat of backdoor attacks. We generate adversarial examples for image augmentation, while for text we employ Easy Data Augmentation (EDA)~\cite{eda}, which includes synonym replacement, random insertion, random swapping, and random deletion. By combining adversarial examples and EDA, we create an augmented dataset to fine-tune the poisoned model, enabling CLIP to defend against backdoor attacks.

\section{Experiments}
\subsection{Experimental Settings}
\textbf{Datasets and CLIP Structure.} In our experiments, we use the 1M data subset of CC3M ~\cite{cc3m} as the training dataset for the CLIP model. We use pre-trained CLIP weights trained on the LAION-400M~\cite{laion} dataset, which is widely used in multilingual image pre-training research ~\cite{a,b,c}. Our results demonstrate that medium-scale network datasets like CC3M are sufficient to train the model effectively while minimizing the need for extensive storage and computing resources. We select the ResNet-50~\cite{resnet} visual encoder for CLIP due to its strong performance in image recognition tasks and its ability to efficiently extract image features. And the text encoder is transformer~\cite{transformer}.

\textbf{Attack Setup.} We employ three different backdoor attack methods: BadNet~\cite{badnet}, Blended~\cite{blended}, and BadCLIP~\cite{badclip}, to poison the CLIP model by fine-tuning it on a subset of the CC3M dataset according to \eqref{eq1}. As illustrated in Fig.\ref{fig}, we introduce a carefully designed trigger into the images to create poisoned image samples and construct sentences containing the target label for the text, forming the poisoned image-text pairs. In the fine-tuning process, we use a 500K subset of the CC3M dataset with a poisoning rate of $0.3\%$. The batch size is $128$, with a total of $10$ batches, and the learning rate is set to $1e-6$, using the cosine scheduling strategy and AdamW~\cite{adam} optimizer. The size of the poisoning patch varies by attack method: BadNet and BadCLIP use a local fixed noise patch of size $16$, while Blended applies a global blending noise. On average, each attack method requires approximately $8$ A100 GPU hours to complete the poisoning process.

\textbf{Defense Setup.} We use the AdvCLIP with our designed backdoor loss function to train on a 250K subset of CC3M data to obtain adversarial samples with backdoor features. We set the magnitude to $0.03$ and use the batch size of $16$. The victim model is based on the CLIP architecture with an ResNet-50 backbone, and training process is conducted over $20$ epochs. After obtaining the adversarial example weights, we fine-tune the poisoned model for defense using the same 250K dataset. We add adversarial perturbations to the original images and perform EDA on the captions to obtain an enhanced dataset. In the defense phase, we set the batch size to $64$, the learning rate to $3e{-}6$, and train the model for $10$ epochs, which takes approximately $14$ A100 GPU hours.

\textbf{Evaluation Metric.} We evaluate the model’s performance using the ImageNet-1K validation dataset. We measure Clean Accuracy (CA) to assess the accuracy of the pre-trained model on clean samples. We calculate the Attack Success Rate (ASR), which indicates the proportion of images with backdoor triggers that are misclassified as the target class by the poisoned model. Our defense algorithm aims to reduce ASR while preserving CA with minimal impact.

\subsection{Main Results}
We conduct experiments on the ImageNet-1K validation dataset, evaluating the defense effectiveness under various backdoor attack methods including BadNet, Blended, and BadCLIP. We successfully verify the effectiveness and superiority of the proposed ABD through comparative experiments with other defense methods, RoCLIP~\cite{roclip} and CleanCLIP~\cite{cleanclip}. The main experimental results are as follows:
\begin{table}[t]
\caption{The performance (\%) of methods on the ImageNet-1K. The best results are shown in bold.}
    \centering
    \resizebox{0.47\textwidth}{!}{
    \begin{tabular}{c|ccccc}
        \toprule
        Attack & Metrics & Victim & Roclip & CleanCLIP & \textbf{Ours(ABD)} \\
        \midrule
        \multirow{2}{*}{BadNet~\cite{badnet}} & CA  & 58.69  & 46.6    & 55.85   & 53.47 \\
                               & ASR & 96.51  & 80.75  & 14.02   & \textbf{5.36} \\
        \multirow{2}{*}{Blended~\cite{blended}} & CA  & 58.48  & 44.55  & 55.53    & 53.29 \\
                               & ASR & 50.28  & 26.96  & 18.25   & \textbf{7.73} \\
        \multirow{2}{*}{BadCLIP~\cite{badclip}} & CA  & 58.62  & 46.47  & 53.98    & 53.4 \\
                                 & ASR & 98.98 & 89.97   & 89.6   & \textbf{35.96} \\
        \bottomrule
    \end{tabular}}
\label{tab1}
\end{table}

As shown in Table \ref{tab1}, our ABD successfully reduces the ASR at the cost of sacrificing minimal CA. Compared to the victim model, the CA of clean samples under the ABD drops by an average of $4.91\%$, while the ASR of backdoor samples decreases by an average of $65.57\%$. This demonstrates the effectiveness of the adversarial examples we designed, indicating their feasibility in mitigating backdoor attacks in multimodal scenario. Furthermore, ABD performs well against various attack methods, including traditional uni-modal attacks like BadNet and Blended, as well as multimodal attacks like BadCLIP, which is specifically designed for CLIP, showcasing ABD's robustness. Compared to the state-of-the-art method CleanCLIP, ABD reduces the ASR by $8.66\%$ for BadNet, $10.52\%$ for Blended, and $53.64\%$ for BadCLIP, with only a $1.73\%$ average decrease in CA.

\subsection{Ablation Study}
To evaluate the impact of adversarial examples in the fine-tuning process, we conduct an ablation experiment focusing on the roles of adversarial examples and text augmentation techniques EDA. We analyze the model's performance with and without adversarial examples, as well as with and without text augmentation by EDA, to determine their individual and combined impacts on defense efficiency. FT represents training without enhancements on images and text, $I_{adv}$ represents training with adversarial samples on images, and $T_{eda}$ represents training with EDA on text.
\begin{table}[t]
\caption{The effects of adversarial patch. The best results are shown in bold.}
    \centering
    \resizebox{0.47\textwidth}{!}{
    \begin{tabular}{c|cccccc}
        \toprule
        Attack & Metrics & Victim & FT & $I_{\text{adv}},T$ & $I,T_{\text{eda}}$ & \textbf{$I_{\text{adv}},T_{\text{eda}}$} \\
        \midrule
        \multirow{2}{*}{BadNet~\cite{badnet}} & CA  & 58.69  & 53.15   & 53.7   & 53.1   & 53.47 \\
                            & ASR & 96.51  & 66.41  & 10.49   & 59.83 &\textbf{5.36} \\
        \multirow{2}{*}{Blended~\cite{blended}} & CA  & 58.48  & 53.68   & 53.68   & 53.72 &53.29 \\
                            & ASR & 50.28  & 53.25  & 18.15  &51.63 & \textbf{7.73} \\
        \multirow{2}{*}{BadCLIP~\cite{badclip}} & CA  & 58.62  & 53.43  & 53.7    & 53.09 &53.4 \\
                            & ASR & 98.98 & 74.92   & 46.03  &75.7 & \textbf{35.96} \\
        \bottomrule
    \end{tabular}}
\label{tab2}
\end{table}
The results shown in Table \ref{tab2} show that fine-tuning with only 250K clean samples does not significantly improve the defense against traditional methods such as BadNet and Blended, as well as the recent BadCLIP attack, leaving a considerable model vulnerability. When adversarial sample enhancement is applied to the image, ASR is reduced by $39.97\%$ on average, and CA is even increased by $0.27\%$. Applying EDA text enhancement alone has a negligible impact on ASR. However, combining adversarial image enhancement with EDA text enhancement leads to further performance improvement, with ASR decreasing by $8.54\%$ and CA decreasing by only $0.3\%$ compared to using adversarial image enhancement alone. The experiments reflect the key role of adversarial examples in defending against CLIP backdoor attacks, and EDA serves as complements. This also reflects the effectiveness and feasibility of bimodal defense.

\begin{table}[t]
\caption{The effects of backdoor loss in Advclip. The best results are shown in bold.}
    \centering
    \resizebox{0.47\textwidth}{!}{
    \begin{tabular}{c|cccc}
        \toprule
        Attack & Metrics & Victim & UAP & \textbf{UAP($\mathcal{L}_{\text{bd}}$)} \\
        \midrule
        \multirow{2}{*}{BadNet~\cite{badnet}} & CA  & 58.69  & 53.29    & 53.7   \\
                               & ASR & 96.51  & 18.39   & \textbf{10.49} \\
        \multirow{2}{*}{Blended~\cite{blended}} & CA  & 58.48 & 54.05    & 53.68 \\
                               & ASR & 50.28  & 30.48   & \textbf{18.15} \\
        \multirow{2}{*}{BadCLIP~\cite{badclip}} & CA  & 58.62  & 53.27      & 53.7 \\
                                 & ASR & 98.98 & 48.3    & \textbf{46.03} \\
        \bottomrule
    \end{tabular}}
\label{tab3}
\end{table}


To investigate the impact of backdoor loss on generating adversarial samples with AdvCLIP, we perform a series of experiments. In Table \ref{tab3}, UAP denotes the adversarial examples generated by the original AdvCLIP model, while UAP($\mathcal{L}_{\text{bd}}$) refers to the improved AdvCLIP model that incorporates the backdoor loss $\mathcal{L}_{\text{bd}}$ , which is shown in \eqref{eq2}. Although the original UAP method provides some defense against backdoor samples, our findings indicate that incorporating the backdoor loss enhances its effectiveness. Specifically, the enhanced AdvCLIP, which accounts for backdoor features, shows a significant improvement in defense performance. The ASR decreases by $7.5\%$, and the CA increases by $0.15\%$, demonstrating a more robust defense with the addition of backdoor loss.
\section{Conclusions}

In this paper, we identify a crucial problem with existing defense methods, such as RoCLIP and CleanCLIP, where the simplicity of their data augmentation strategies results in suboptimal performance in mitigating backdoor attacks. To address this limitation, we propose a novel approach called Adversarial Backdoor Defense (ABD) within the CLIP. Our method generates adversarial examples that closely approximate backdoor samples in the feature space, leveraging them as an effective data augmentation technique for backdoor defense. Furthermore, we uncover subtle connections between adversarial examples and backdoor samples in compromised models. Through extensive experiments on the ImageNet-1K validation dataset, we demonstrate that ABD consistently outperforms existing methods, including RoCLIP and CleanCLIP. It should be noted that ABD significantly reduces the success rate of backdoor attacks with only a minimal impact on clean accuracy. Our research provides a valuable insight into the application of adversarial examples in CLIP and the revealing of the underlying mechanism of the correlation between adversarial and backdoor samples.


\vspace{12pt}


\newpage
\begin{thebibliography}{00}
\bibitem{rl}
Y. Bengio, A. Courville, and P. Vincent, ``Representation learning: A review and new perspectives,'' \emph{IEEE Transactions on Pattern Analysis and Machine Intelligence}, pp. 1798--1828, 2013.

\bibitem{clip}
A. Radford, J. W. Kim, C. Hallacy, A. Ramesh, G. Goh, S. Agarwal, G. Sastry, A. Askell, P. Mishkin, J. Clark, \emph{et al.}, ``Learning transferable visual models from natural language supervision,'' in \emph{International Conference on Machine Learning}, PMLR, pp. 8748--8763, 2021.

\bibitem{align}
C. Jia, Y. Yang, Y. Xia, Y.-T. Chen, Z. Parekh, H. Pham, Q. Le, Y.-H. Sung, Z. Li, and T. Duerig, "Scaling up visual and vision-language representation learning with noisy text supervision," in \textit{Proc. Int. Conf. Machine Learning}, pp. 4904--4916, 2021.

\bibitem{basic}
H. Pham, Z. Dai, G. Ghiasi, K. Kawaguchi, H. Liu, A. W. Yu, J. Yu, Y.-T. Chen, M.-T. Luong, Y. Wu, et al., "Combined scaling for open-vocabulary image classification," \textit{arXiv preprint arXiv:2111.10050}, p. 4, 2021.

\bibitem{bcl} 
N. Carlini and A. Terzis, ``Poisoning and Backdooring Contrastive Learning,'' in \emph{International Conference on Learning Representations}, 2022.


\bibitem{pwb}
N. Carlini, M. Jagielski, C. A. Choquette-Choo, D. Paleka, W. Pearce, H. Anderson, A. Terzis, K. Thomas, and F. Tram\`er, ``Poisoning Web-Scale Training Datasets is Practical,'' in \emph{2024 IEEE Symposium on Security and Privacy}, IEEE Computer Society, pp. 176--176, 2024.

\bibitem{tijo}
I. Sur, K. Sikka, M. Walmer, K. Koneripalli, A. Roy, X. Lin, A. Divakaran, and S. Jha, ``TIJO: Trigger Inversion with Joint Optimization for Defending Multimodal Backdoored Models,'' \emph{IEEE International Conference on Computer Vision}, pp. 165-175, 2023.

\bibitem{baden}
J. Jia, Y. Liu, and N. Gong, ``BadEncoder: Backdoor Attacks to Pre-trained Encoders in Self-Supervised Learning,'' \emph{IEEE Symposium on Security and Privacy}, pp. 2043-2059, 2022.

\bibitem{baibadclip}
J. Bai, K. Gao, S. Min, S. Xia, Z. Li, and W. Liu, ``BadCLIP: Trigger-Aware Prompt Learning for Backdoor Attacks on CLIP,'' \emph{IEEE/CVF Conference on Computer Vision and Pattern Recognition}, 2024.


\bibitem{pla} W. Wang, C. Du, T. Wang, K. Zhang, W. Luo, L. Ma, W. Liu, and X. Cao, ``Punctuation-level Attack: Single-shot and Single Punctuation Attack Can Fool Text Models,'' \emph{Proceedings of the 37th International Conference on Neural Information Processing Systems}, pp.~49312--49324, 2023.

\bibitem{nad}
Y. Li, X. Lyu, N. Koren, L. Lyu, B. Li, and X. Ma, ``Neural Attention Distillation: Erasing Backdoor Triggers from Deep Neural Networks,'' \emph{International Conference on Learning Representations}, 2021.

\bibitem{detbd}
S. Feng, G. Tao, S. Cheng, G. Shen, X. Xu, Y. Liu, K. Zhang, S. Ma, and X. Zhang, ``Detecting Backdoors in Pre-trained Encoders,'' \emph{Computer Vision and Pattern Recognition}, pp. 16352-16362, 2023.

\bibitem{abl}
Y. Li, X. Lyu, N. Koren, L. Lyu, B. Li, and X. Ma, ``Anti-Backdoor Learning: Training Clean Models on Poisoned Data,'' \emph{Conference on Neural Information Processing Systems}, pp. 14900-14912, 2021.

\bibitem{roclip}
W. Yang, J. Gao, and B. Mirzasoleiman, ``Robust contrastive language-image pretraining against data poisoning and backdoor attacks,'' \emph{Advances in Neural Information Processing Systems}, 2024.

\bibitem{cleanclip}
H. Bansal, N. Singhi, Y. Yang, F. Yin, A. Grover, and K.-W. Chang, ``Cleanclip: Mitigating data poisoning attacks in multimodal contrastive learning,'' in \emph{Proceedings of the IEEE/CVF International Conference on Computer Vision}, 2023, pp. 112--123.

\bibitem{adv1}Gao, Y., Wu, D., Zhang, J., Gan, G., Xia, S., Niu, G. \& Sugiyama, M. On the Effectiveness of Adversarial Training Against Backdoor Attacks. {\em IEEE Transactions On Neural Networks And Learning Systems}, pp. 1-11, 2024. 

\bibitem{adv4}
M. Xue, Y. Wu, Z. Wu, Y. Zhang, J. Wang, and W. Liu, ``Detecting backdoor in deep neural networks via intentional adversarial perturbations,'' \emph{Information Sciences}, vol. 634, pp. 564--577, 2023.

\bibitem{adv5}
S. Wei, M. Zhang, H. Zha, and B. Wu, ``Shared adversarial unlearning: Backdoor mitigation by unlearning shared adversarial examples,'' \emph{Conference on Neural Information Processing Systems}, 2023.

\bibitem{adv2}Weng, C., Lee, Y. \& Wu, S. On the Trade-off between Adversarial and Backdoor Robustness.. {\em Conference On Neural Information Processing Systems}. 2020.

\bibitem{adv6}
Y. Zeng, S. Chen, W. Park, Z. Mao, M. Jin, and R. Jia, ``Adversarial unlearning of backdoors via implicit hypergradient,'' \emph{International Conference on Learning Representations}, 2022.

\bibitem{adv7}
D. Wu and Y. Wang, ``Adversarial neuron pruning purifies backdoored deep models,'' \emph{Conference on Neural Information Processing Systems}, pp. 16913--16925, 2021.

\bibitem{adv3}Gan, Z., Chen, Y., Li, L., Zhu, C., Cheng, Y. \& Liu, J. Large-Scale Adversarial Training for Vision-and-Language Representation Learning.. {\em Conference On Neural Information Processing Systems}. 2020.

\bibitem{adv8}
P.-F. Zhang, Z. Huang, and G. Bai, ``Universal adversarial perturbations for vision-language pre-trained models,'' \emph{Proceedings of the 47th International ACM SIGIR Conference on Research and Development in Information Retrieval}, pp. 862--871, 2024.

\bibitem{adv9}
A. Madry, A. Makelov, L. Schmidt, D. Tsipras, and A. Vladu, ``Towards deep learning models resistant to adversarial attacks,'' \emph{International Conference on Learning Representations}, abs/1706.06083, 2017.

\bibitem{imagenet}
J. Deng, W. Dong, R. Socher, L.-J. Li, K. Li, and L. Fei-Fei, ``Imagenet: A large-scale hierarchical image database,'' in \emph{2009 IEEE Conference on Computer Vision and Pattern Recognition}, pp. 248--255, 2009.

\bibitem{pbe}
B. Mu, Z. Niu, L. Wang, X. Wang, Q. Miao, R. Jin, and G. Hua, ``Progressive backdoor erasing via connecting backdoor and adversarial attacks,'' in \emph{Proceedings of the IEEE/CVF Conference on Computer Vision and Pattern Recognition}, pp. 20495--20503, 2023.

\bibitem{tur}
Z. Niu, Y. Sun, Q. Miao, R. Jin, and G. Hua, ``Towards unified robustness against both backdoor and adversarial attacks,'' \emph{IEEE Transactions on Pattern Analysis and Machine Intelligence}, pp. 1--16, 2024.

\bibitem{advclip}
Z. Zhou, S. Hu, M. Li, H. Zhang, Y. Zhang, and H. Jin, ``Advclip: Downstream-agnostic adversarial examples in multimodal contrastive learning,'' in \emph{Proceedings of the 31st ACM International Conference on Multimedia}, pp. 6311--6320, 2023.

\bibitem{gan}
I. Goodfellow, J. Pouget-Abadie, M. Mirza, B. Xu, D. Warde-Farley, S. Ozair, A. Courville, and Y. Bengio, ``Generative adversarial networks,'' \emph{Communications of the ACM}, pp. 139--144, 2020.

\bibitem{infonce}
A. van den Oord, Y. Li, and O. Vinyals, ``Representation learning with contrastive predictive coding,'' \emph{arXiv preprint arXiv:1807.03748}, 2018.

\bibitem{eda}
J. Wei and K. Zou, ``EDA: Easy Data Augmentation Techniques for Boosting Performance on Text Classification Tasks,'' in \emph{Proceedings of the 2019 Conference on Empirical Methods in Natural Language Processing and the 9th International Joint Conference on Natural Language Processing}, pp. 6382--6388, 2019.

\bibitem{cc3m}
P. Sharma, N. Ding, S. Goodman, and R. Soricut, ``Conceptual captions: A cleaned, hypernymed, image alt-text dataset for automatic image captioning,'' in \emph{Proceedings of the 56th Annual Meeting of the Association for Computational Linguistics}, pp. 2556--2565, 2018.

\bibitem{laion}
C. Schuhmann, R. Vencu, R. Beaumont, R. Kaczmarczyk, C. Mullis, A. Katta, T. Coombes, J. Jitsev, and A. Komatsuzaki, ``Laion-400m: Open dataset of clip-filtered 400 million image-text pairs,'' \emph{arXiv preprint arXiv:2111.02114}, 2021.

\bibitem{a}
Y. Li, F. Liang, L. Zhao, Y. Cui, W. Ouyang, J. Shao, F. Yu, and J. Yan, ``Supervision Exists Everywhere: A Data Efficient Contrastive Language-Image Pre-training Paradigm,'' in \emph{Proceedings of the International Conference on Learning Representations}, 2024.

\bibitem{b}
N. Mu, A. Kirillov, D. Wagner, and S. Xie, ``SLIP: Self-Supervision Meets Language-Image Pre-Training,'' in \emph{Proceedings of the European Conference on Computer Vision}, pp. 529--544, 2022.

\bibitem{c}
S. Goel, H. Bansal, S. Bhatia, R. Rossi, V. Vinay, and A. Grover, ``CyCLIP: Cyclic Contrastive Language-Image Pretraining,'' \emph{Advances in Neural Information Processing Systems}, pp. 6704--6719, 2022.

\bibitem{resnet}
B. Koonce, ``ResNet 50,'' \emph{Convolutional Neural Networks with Swift for TensorFlow: Image Recognition and Dataset Categorization}, pp. 63--72, 2021.

\bibitem{transformer}
K. Han, A. Xiao, E. Wu, J. Guo, C. Xu, and Y. Wang, ``Transformer in transformer,'' \emph{Advances in Neural Information Processing Systems}, pp. 15908--15919, 2021.

\bibitem{badnet}
T. Gu, B. Dolan-Gavitt, and S. Garg, ``Badnets: Identifying vulnerabilities in the machine learning model supply chain,'' \emph{arXiv preprint arXiv:1708.06733}, 2017.

\bibitem{blended}
X. Chen, C. Liu, B. Li, K. Lu, and D. Song, ``Targeted backdoor attacks on deep learning systems using data poisoning,'' \emph{arXiv preprint arXiv:1712.05526}, 2017.

\bibitem{badclip}
S. Liang, M. Zhu, A. Liu, B. Wu, X. Cao, and E.-C. Chang, ``Badclip: Dual-embedding guided backdoor attack on multimodal contrastive learning,'' in \emph{Proceedings of the IEEE/CVF Conference on Computer Vision and Pattern Recognition}, pp. 24645--24654, 2024.

\bibitem{adam}
I. Loshchilov, ``Decoupled weight decay regularization,'' \emph{arXiv preprint arXiv:1711.05101}, 2017.

\end{thebibliography}
\end{document}